\documentclass[sigconf]{acmart}

\AtBeginDocument{%
  }

\setcopyright{acmlicensed}
\copyrightyear{2026}
\acmYear{2026}
\acmDOI{XXXXXXX.XXXXXXX}
\acmConference[MFMB '26]{ICAIF Workshop on Multimodal and Foundation Models in Banking}{November 15, 2026}{Milan, Italy}
\acmISBN{978-1-4503-XXXX-X/2018/06}

\acmSubmissionID{---}

\newcommand{\Expectover}[2]{\mathbb{E}_{#1}\!\left[#2\right]}

\usepackage[most]{tcolorbox}
\newtcolorbox{promptbox}{
  enhanced,
  breakable,
  colback=gray!3,
  colframe=gray!50,
  boxrule=0.5pt,
  arc=1mm,
  left=6pt,right=6pt,top=6pt,bottom=6pt,
  listing only,
  listing options={
    basicstyle=\ttfamily\small,
    breaklines=true,
    columns=fullflexible
  }
}
\usepackage{graphicx}

\usepackage[table]{xcolor}
\begin{document}

\title{Multimodal Zero-Shot Prediction for Transaction Data at Scale}
%\titlenote{An abridged version of this paper was presented at the ICAIF'26 Workshop on Multimodal and Foundation Models in Banking.}

% \begingroup 
% \renewcommand{\thefootnote}{} 
% \titlenote{An abridged version of this paper was presented at the ICAIF'26 Workshop on Multimodal and Foundation Models in Banking.} 
% \endgroup

% \begingroup \renewcommand\thefootnote{} \footnotetext{An abridged version of this paper was presented at the ICAIF'26 Workshop on Multimodal and Foundation Models in Banking.} \addtocounter{footnote}{-1} \endgroup

\author{Parameswaran Kamalaruban}
\authornote{Both authors contributed equally to this research.}
\affiliation{%
  \institution{Risk and Security AI Lab, Visa Inc.}
  \country{United Kingdom}
}
\email{kaparame@visa.com}
%\orcid{xxxx-xxxx-xxxx}
%\correspondingauthor

\author{Viktor Drobnyi}
\authornotemark[1]
\affiliation{%
  \institution{Risk and Security AI Lab, Visa Inc.}
  \country{United Kingdom}
}
\email{vdrobnyi@visa.com}

\author{Maeve Madigan}
\affiliation{%
  \institution{Risk and Security AI Lab, Visa Inc.}
  \country{United Kingdom}
}
\email{mmadigan@visa.com}

\author{Julia Rozanova}
\affiliation{%
  \institution{Risk and Security AI Lab, Visa Inc.}
  \country{United Kingdom}
}
\email{yrozanov@visa.com}

\author{David Sutton}
\affiliation{%
  \institution{Risk and Security AI Lab, Visa Inc.}
  \country{United Kingdom}
}
\email{dsutton@visa.com}

\author{Stuart Burrell}
\affiliation{%
  \institution{Risk and Security AI Lab, Visa Inc.}
  \country{United Kingdom}
}
\email{sburrell@visa.com}

\renewcommand{\shortauthors}{Kamalaruban et al.}

% !TEX root = main.tex
%%%%%%%%%%%%%%%%%%%%%%%%%%%%%%%%%%%%%
%%%%%%%%%%%%%%%%%%%%%%%%%%%%%%%%%%%%%
\begin{abstract}
Sequential financial transaction data are increasingly being used by financial institutions to train payment foundation models at scale, producing rich contextual embeddings that support applications such as fraud prevention, credit risk assessment, and offer personalization. However, such models are not designed for flexible instruction-following reasoning and require bespoke downstream models for each application. In contrast, large language models (LLMs) provide a natural-language interface for flexible transaction reasoning, yet naively serializing transaction histories into text representations is inefficient and costly. Multimodal fusion architectures bridge these paradigms, injecting transaction embeddings directly into flexible language reasoning pipelines. In this work, we present a systematic study of how transaction embeddings, pretrained at the scale of a global payments network, interact with LLM-based reasoning through embedding injection, transaction-language alignment, and instruction tuning. To do so, we train the Multimodal Instruction Network for Transactions (MINT), which combines a transaction encoder pretrained on billions of real-world payments with multimodal instruction tuning on millions of transaction question-answering examples. We evaluate predictive and extractive question answering, zero-shot generalization to unseen question types, and inference efficiency. Our results reveal a consistent trade-off: compact transaction embeddings improve predictive QA accuracy while substantially reducing input tokens, latency, and memory relative to a text-serialized baseline, whereas serialized histories remain highly competitive for extractive QA. Overall, our empirical findings provide a set of actionable insights and practical guidance for building multimodal transaction-aware language systems at network scale.
\begingroup 
\renewcommand{\thefootnote}{} 
\footnote{An abridged version of this paper was presented at the ICAIF'26 Workshop on Multimodal and Foundation Models in Banking.} 
\endgroup
\end{abstract}



%%%%%%%%%%%%%% COMMENT (ARXIV) %%%%%%%%%%%%%%
%%%%%%%%%%%%%% COMMENT (ARXIV) %%%%%%%%%%%%%%

\maketitle

% !TEX root = main.tex
%%%%%%%%%%%%%%%%%%%%%%%%%%%%%%%%%%%%%
%%%%%%%%%%%%%%%%%%%%%%%%%%%%%%%%%%%%%
\vspace{-2mm}
\section{Introduction}
\label{sec:introduction}

\begin{figure}[ht]
    \centering
    \includegraphics[width=\linewidth]{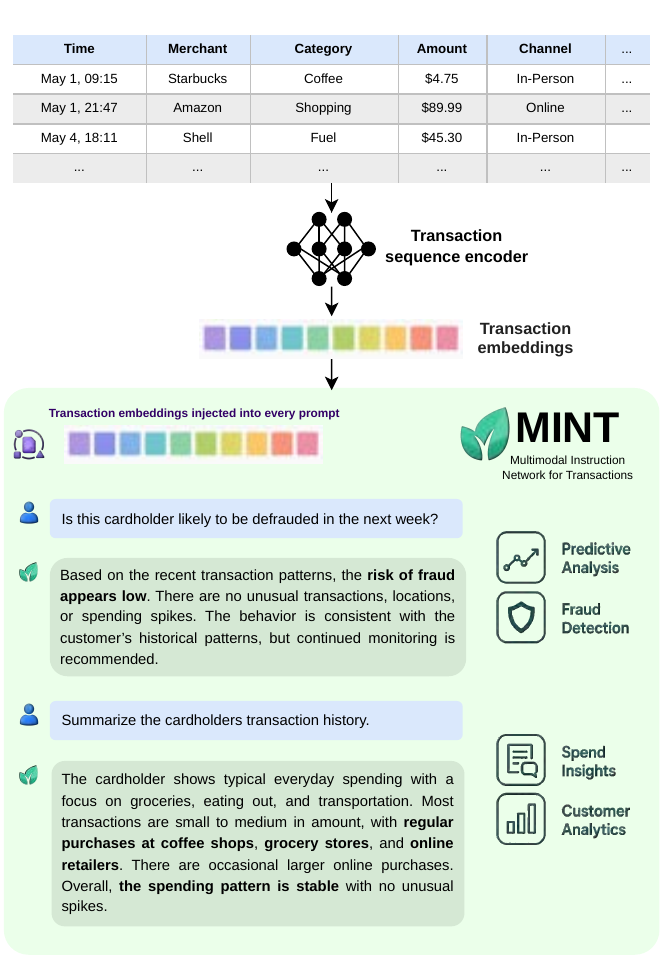}
    \caption{Overview of MINT. Transaction histories are encoded into compact transaction embeddings, enabling predictive question answering and financial reasoning through a natural-language interface. Example applications include forecasting future customer behavior, such as transaction volume and average transaction amount over the next 30 days, alongside summarization, fraud-risk assessment, and financial analysis.}
    \label{fig:txn-reasoner-use-cases}
\end{figure}

\begin{figure}[t]
    \centering
    \vspace{1.5em}
    \includegraphics[width=\linewidth]{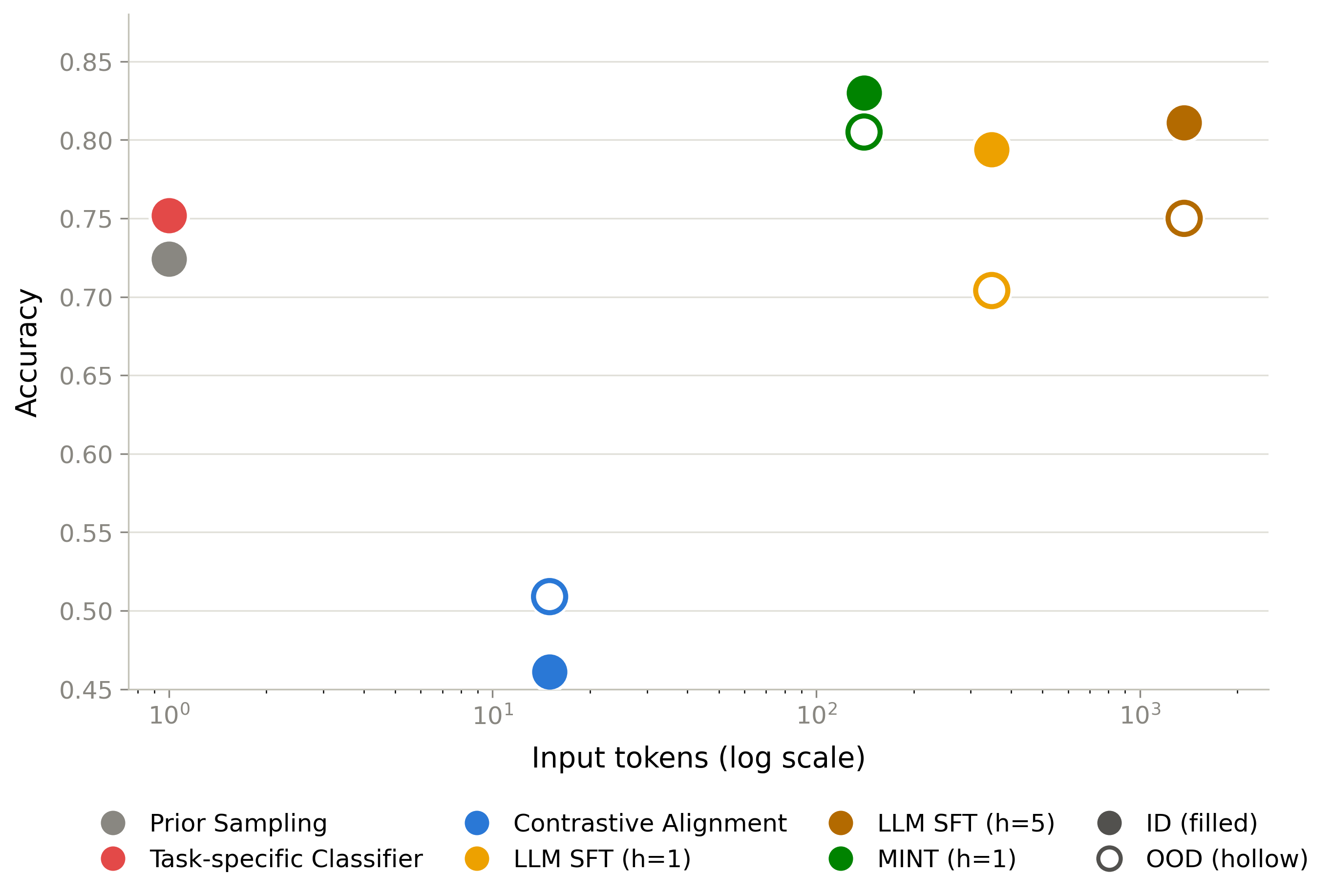}
    \caption{Accuracy-efficiency trade-off on predictive QA. The x-axis shows input token count and the y-axis shows accuracy; filled and unfilled markers correspond to in-distribution (ID) and out-of-distribution (OOD) evaluation, respectively. MINT ($h=1$) occupies a more favorable region than the text-serialized baseline (LLM SFT), balancing accuracy with inference efficiency.}
    \label{fig:pareto-plot}
\end{figure}

Financial behavior often manifests itself as a sequence of discrete events, such as a temporally ordered history of transactions, where each transaction comprises heterogeneous categorical and numerical attributes (e.g., merchant, amount, channel, and timestamp). Transaction understanding is central to applications such as fraud detection, risk assessment, and customer analytics, and reasoning over such data requires modeling both interactions among fields within a transaction and temporal dependencies across transactions, often over long horizons. Self-supervised pretraining on large-scale transaction corpora has produced increasingly powerful foundation models~\citep{skalski2023towards,li2025panther,braithwaite2025your,aminian2025fraudtransformer,dou2025transactiongpt,yeh2026treasure,ostroukhov2026pragma}. These encoders learn transferable representations from behavioral sequences and achieve strong performance across a wide range of downstream financial tasks.

Large language models (LLMs) provide a compelling interface for transaction understanding due to their natural-language interaction and instruction-following capabilities. A straightforward approach serializes transaction histories into text, but this scales poorly with sequence length and often fails to exploit the structured numerical and temporal information present in transaction data~\citep{gruver2023large,kong2025time,zhang2025timemaster,luo2025time}. Recent work on event-sequence and time-series reasoning instead combines a specialized sequence encoder with an LLM, conditioning on compact learned representations rather than lengthy textual serializations~\citep{chow2024towards,xie2024chatts,langer2025opentslm,yu2025ts,abdullaeva2024esqa,raman2024scalable}. \citet{yu2025ts} align a time-series foundation model with LLM reasoning for \emph{unimodal} time series, while \citet{abdullaeva2024esqa} and \citet{raman2024scalable} apply related ideas to financial event sequences, typically through joint encoder-LLM training or task-specific fusion architectures beyond a simple projector.

In this work, we present a systematic study of how transaction embeddings, pretrained at the scale of a global payments network, interact with LLM-based reasoning. To do so, we train the Multimodal Instruction Network for Transactions (\textsc{MINT}), which couples a transaction foundation model pretrained on billions of real-world payment events to a decoder-only LLM through lightweight embedding injection, transaction-language alignment, and LoRA-based instruction tuning (see Figure~\ref{fig:txn-reasoner-use-cases}). Each of these ingredients has precedent in the vision-language and time-series-LLM literature; \textsc{MINT} does not introduce a new multimodal architecture. Our contribution is instead to combine these established components at real-world payments-network scale, and to characterize their behavior across predictive and extractive question answering, previously unseen question types, training-data composition, and inference efficiency (see Figure~\ref{fig:pareto-plot}). To our knowledge, this is the first empirical study of this kind on a proprietary transaction corpus at this scale, and it yields a set of actionable insights and practical guidance for training multimodal transaction-aware language systems.

% Our main contributions are: 
% \begin{itemize} 
% \item \textbf{Introduce} \textsc{MINT}, a multimodal transaction reasoning system that couples a large-scale pretrained, frozen transaction encoder with a decoder-only LLM via embedding injection, modality alignment, and parameter-efficient adaptation.  
% \item \textbf{Validate} that compact transaction embeddings improve predictive-QA accuracy and inference efficiency (input tokens, latency, memory) relative to a text-serialized baseline trained with the same recipe, while text serialization remains competitive for extractive QA (see Figure~\ref{fig:pareto-plot}). 
% \item \textbf{Present} a comprehensive study of multimodal transaction reasoning, analyzing representation layer, model capacity, training-data composition, history length, robustness, and efficiency, yielding practical design guidance for transaction-language systems. 
% \end{itemize}

% !TEX root = main.tex
%%%%%%%%%%%%%%%%%%%%%%%%%%%%%%%%%%%%%
%%%%%%%%%%%%%%%%%%%%%%%%%%%%%%%%%%%%%
\vspace{-2mm}
\section{Related Work}
\label{sec:related-work}

\textbf{Event sequence foundation models.} Transaction histories can be represented as structured event sequences with heterogeneous attributes and timestamps. Prior work on tabular and event-sequence modeling has shown the importance of hierarchical architectures that separately encode field-level information within events and temporal dependencies across events~\citep{padhi2021tabular,Zhang2023CrossformerTU,Azorin2024FinegrainedAI}. More recently, large-scale self-supervised pretraining on transaction corpora has established pretrained transaction sequence encoders as strong foundation models for behavioral data~\citep{skalski2023towards,li2025panther,braithwaite2025your,aminian2025fraudtransformer,dou2025transactiongpt,yeh2026treasure,ostroukhov2026pragma}. In particular, \citet{skalski2023towards} showed that generative autoregressive pretraining yields highly effective transaction representations, outperforming both traditional feature-engineering approaches and alternative self-supervised objectives for transaction modeling~\citep{jha2012employing,clark2020electra,babaev2022coles}. Their results demonstrated that autoregressively pretrained transaction encoders transfer well across a diverse range of downstream tasks. This motivates our use of a pretrained transaction sequence encoder and makes the corresponding task-specific classifier a strong baseline throughout our evaluation.

\textbf{Multimodal event sequence reasoning.} Recent work has explored adapting LLMs to time-series and event-sequence reasoning, highlighting challenges in long-context processing, numerical reasoning, and faithful generation~\citep{gruver2023large,kong2025time,zhang2025timemaster,luo2025time}. To address these limitations, multimodal approaches combine specialized sequence encoders with LLMs through projection layers, cross-attention modules, or other fusion mechanisms~\citep{chow2024towards,xie2024chatts,langer2025opentslm,yu2025ts}. Transaction-focused reasoning systems extend this paradigm to financial event sequences~\citep{abdullaeva2024esqa,raman2024scalable}. However, existing approaches typically omit one or more ingredients that have become standard in modern multimodal reasoning systems, including explicit modality-alignment training, chain-of-thought supervision, and comprehensive evaluation across both in-distribution and out-of-distribution settings. 

A key distinction of our approach is the decoupling of transaction representation learning from reasoning-model training. Prior work either jointly optimizes the transaction sequence encoder, connector, and LLM~\citep{abdullaeva2024esqa} or employs more complex fusion architectures~\citep{raman2024scalable}. In contrast, following contemporary vision-language model training practices~\citep{liu2023visual,marafioti2025smolvlm}, we first pretrain a transaction sequence encoder on large-scale transaction data, then freeze it during modality alignment and instruction tuning. This separation substantially simplifies training, allows transaction representations to be learned efficiently without repeatedly processing massive datasets through the LLM, and enables the use of a lightweight MLP projector rather than more complex architectures such as Q-Former~\citep{li2023blip}. Furthermore, unlike approaches that rely on task identifiers, task embeddings, or task-specific control tokens~\citep{abdullaeva2024esqa,raman2024scalable}, our formulation treats transaction understanding as a unified instruction-following problem. As a result, the model is not tied to a fixed set of task definitions and can naturally generalize to previously unseen question types and instructions. Beyond predictive accuracy, we also provide a comprehensive evaluation spanning robustness, efficiency, latency, memory consumption, and detailed ablations of the major design choices.

\textbf{Alignment of event-sequence and text representations.} A complementary line of work aligns event-sequence representations with semantic text embeddings using contrastive objectives~\citep{fadeev2025latte,zhang2025sensorlm}. These methods primarily aim to learn transferable representations for downstream predictive tasks. In contrast, we use transaction-language alignment as an intermediate stage in a multimodal reasoning training pipeline, ultimately targeting generative tasks such as transaction captioning and question answering.
% !TEX root = main.tex
%%%%%%%%%%%%%%%%%%%%%%%%%%%%%%%%%%%%%
%%%%%%%%%%%%%%%%%%%%%%%%%%%%%%%%%%%%%
\vspace{-2mm}
\section{Model Architecture}
\label{sec:model-architecture}

\begin{figure}[ht]
    \centering
    \includegraphics[width=\linewidth]{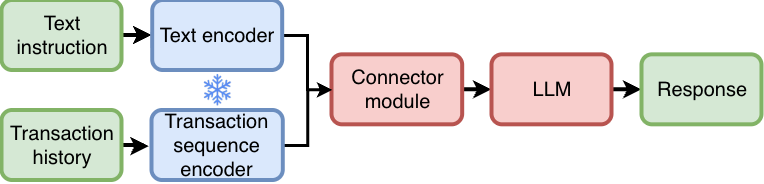}
    \caption{\textsc{MINT} architecture: frozen transaction sequence encoder produces per-transaction embeddings, a trainable connector projects them into the LLM hidden space, and a decoder-only LLM adapted via LoRA generates text outputs conditioned on injected embeddings.}
    \label{fig:txn-reasoner-model-arch}
\end{figure}

\textsc{MINT} comprises three components: a \textbf{transaction sequence encoder} producing transaction sequence embeddings, a \textbf{trainable connector} mapping these embeddings into the LLM hidden space, and a \textbf{decoder-only LLM} adapted via LoRA~\cite{hu2022lora}.

\textbf{Transaction sequence encoder.} Let $\mathbf{T}=\{(x_i,t_i)\}_{i=1}^{N}$ denote a customer's transaction history, where $x_i$ is a transaction record and $t_i$ its timestamp. Following prior work on tabular time-series modeling~\citep{padhi2021tabular,skalski2023towards}, we use a Transformer-based transaction sequence encoder. Numerical attributes are log-transformed, categorical attributes are embedded via field-specific lookup tables, and a field-level Transformer models interactions within each transaction. A sequence-level Transformer then captures temporal dependencies across transactions, producing contextualized embeddings $e_i = E(x_i \mid \{(x_j,t_j)\}_{j=1}^{i}) \in \mathbb{R}^{d_e}$. The sequence-level Transformer operates over windows of up to 128 transactions using sliding-window attention with a window size of 64. The encoder is pretrained with an autoregressive next-event objective, $\mathcal{L}_{\text{AR}} = -\sum_{i=1}^{N} \log p(x_i \mid x_{<i})$, and remains frozen during multimodal training, serving solely as a transaction feature extractor.

\textbf{Connector and LLM backbone.} A connector $g_\phi$ (parameterized by $\phi$) maps the transaction embedding $e_i\in\mathbb{R}^{d_e}$ to the LLM hidden space $\mathbb{R}^d$: $z_i=g_\phi(e_i)\in\mathbb{R}^d$. We use a multi-layer perceptron (MLP) with normalization: $g_\phi(e) = \mathrm{LN}\!\left(W_2\,\sigma\!\left(W_1\,\mathrm{LN}(e)\right)\right)$, where $\sigma$ is GELU, LN denotes layer norm, $W_1\in\mathbb{R}^{\ell\times d_e}$, $W_2\in\mathbb{R}^{d\times \ell}$, and $\ell$ is the connector hidden size. The LLM backbone is an instruction-tuned decoder-only model adapted via LoRA with frozen base weights $\theta$, so gradients flow only through LoRA parameters $\Delta$ and connector parameters $\phi$, preserving general linguistic competence while enabling parameter-efficient specialization.

\textbf{Embedding injection.} Transaction histories are injected by replacing \texttt{<emb>} placeholder tokens in the prompt with the corresponding projected embeddings $z_i$, so the LLM conditions on a fixed-size representation of the history rather than a text serialization whose length grows with history size.

% !TEX root = main.tex
%%%%%%%%%%%%%%%%%%%%%%%%%%%%%%%%%%%%%
%%%%%%%%%%%%%%%%%%%%%%%%%%%%%%%%%%%%%
\vspace{-2mm}
\section{Datasets}
\label{sec:datasets}

We formulate transaction understanding as a unified instruction-following problem that combines transaction history embeddings with natural-language instructions. The model is trained on two task families: \textbf{transaction captioning}, which generates summaries of customer behavior (e.g., spending patterns, merchant preferences, and temporal trends), and \textbf{transaction question answering (QA)}, which answers questions grounded in transaction history. We focus on \emph{multiple-choice} QA, including both \emph{extractive} questions, where the correct answer is directly supported by observed transactions, and \emph{predictive} questions, which require inference or forecasting from historical behavior. Depending on the task, the model may additionally generate natural-language rationales alongside the selected answer.

\textbf{Transaction data.} We use a proprietary dataset of anonymized card transactions collected from a real-world payment network. Each transaction record contains a customer identifier, timestamp, and a set of structured attributes comprising numerical features (e.g., transaction amount) and categorical features (e.g., merchant identity, merchant category, and location). Due to the sensitive nature of the data, specific attribute names cannot be disclosed. The transaction sequence encoder is pretrained on a large corpus spanning billions of transactions. For multimodal instruction tuning and evaluation, we use a \emph{separate}, customer-level 80/10/10 train/validation/test split (134k/14k/19k transactions), with no customer overlap across splits. This instruction-tuning transaction dataset serves as the source data from which all captioning and QA examples are generated. It is isolated from the pretraining corpus at the customer level, and all transactions are timestamped strictly after the pretraining cut-off date to prevent temporal leakage.

\textbf{Caption and QA data.} Captions combine template-generated structured reports with LLM-synthesized semantic descriptions (134k/14k/19k train/validation/test examples). We use Mistral-7B-Instruct-v0.3~\cite{mistral7b-instruct-v03} as the teacher LLM. QA data are generated programmatically from transaction attributes and temporal aggregates with deterministic ground-truth answers, yielding 5.74M/605k/813k extractive QA examples (43 question types) and 3.47M/366k/492k predictive QA examples (26 question types). To evaluate OOD generalization, we additionally construct test sets containing previously \emph{unseen question types}: 170k extractive examples (9 types) and 189k predictive examples (10 types). Chain-of-thought (CoT) rationales are synthesized using the same teacher LLM conditioned on the transaction history, question, and ground-truth answer, then filtered using answer-consistency and length constraints, resulting in 94k/16k/26k extractive and 362k/50k/80k predictive examples.

% !TEX root = main.tex
%%%%%%%%%%%%%%%%%%%%%%%%%%%%%%%%%%%%%
%%%%%%%%%%%%%%%%%%%%%%%%%%%%%%%%%%%%%
\vspace{-2mm}
\section{Training}
\label{sec:training}

%We train the model in two stages. \textbf{Modality alignment} trains only the connector $g_\phi$ on captioning data $\mathcal{D}_{\text{cap}}$ with a completion-only autoregressive loss over response tokens, with the base LLM frozen and LoRA disabled. \textbf{Supervised fine-tuning (SFT)} then jointly optimizes $\phi$ and LoRA parameters $\Delta$ on a mixture $\mathcal{D} = \mathcal{D}_{\text{cap}} \cup \mathcal{D}_{\text{qa}}$ of captioning and QA data (with and without CoT rationales), again with a completion-only autoregressive loss. 

We train the model in three stages. \textbf{Encoder pretraining} first trains the transaction encoder independently on large-scale transaction data to produce fixed transaction embeddings. \textbf{Modality alignment} then trains only the connector $g_\phi$ on captioning data $\mathcal{D}_{\text{cap}}$ with a completion-only autoregressive loss over response tokens, with the base LLM frozen and LoRA disabled. \textbf{Supervised fine-tuning (SFT)} finally jointly optimizes $\phi$ and LoRA parameters $\Delta$ on a mixture $\mathcal{D} = \mathcal{D}_{\text{cap}} \cup \mathcal{D}_{\text{qa}}$ of captioning and QA data (with and without CoT rationales), again with a completion-only autoregressive loss.

%The decoupling of transaction representation learning (encoder pretraining) from reasoning-model training (connector and LoRA) follows the standard vision-language model training recipe~\citep{liu2023visual,marafioti2025smolvlm}: it avoids repeatedly routing large transaction corpora through the LLM, and allows pretrained transaction representations to be aligned with the LLM using only a few hundred thousand captioning examples, after which multimodal instruction tuning is performed on approximately 10 million transaction QA examples.

The separation of encoder pretraining from downstream multimodal training (modality alignment and SFT) follows the standard vision-language model recipe~\citep{liu2023visual,marafioti2025smolvlm}: it avoids repeatedly routing large transaction corpora through the LLM and allows pretrained transaction representations to be aligned with the LLM using only a few hundred thousand captioning examples, after which multimodal instruction tuning is performed on samples drawn from a pool of approximately 10 million transaction QA examples.

%it avoids repeatedly routing large transaction corpora through the LLM, and allows pretrained transaction representations to be aligned with the LLM using only a few hundred thousand supervised captioning and QA examples.

% !TEX root = main.tex
%%%%%%%%%%%%%%%%%%%%%%%%%%%%%%%%%%%%%
%%%%%%%%%%%%%%%%%%%%%%%%%%%%%%%%%%%%%
\vspace{-2mm}
\section{Experiments}
\label{sec:experiments}

%%%%%%%%%%%%%%%%%%%%%%%%%%%%%%%%%%%%% Table %%%%%%%%%%%%%%%%%%%%%%%%%%%%%%%%%%%%%
\begin{table*}[t]
\caption{Overall accuracy on extractive and predictive QA, with inference efficiency metrics (TTFT, decode throughput, peak VRAM, and input token count) measured on predictive QA. Results marked with $^{\dagger}$ are trained and evaluated on the same question set and are included for reference only. Best values are highlighted in bold, and second-best values are underlined.}
\label{tab:qa_main}
\centering
\small
\resizebox{0.86\textwidth}{!}{
\begin{tabular}{lrrrr rrrr} 
\toprule 
& \multicolumn{2}{c}{\textbf{Predictive QA}} & \multicolumn{2}{c}{\textbf{Extractive QA}} & \multicolumn{4}{c}{\textbf{Inference Metrics}} \\ 
\cmidrule(lr){2-3}\cmidrule(lr){4-5}\cmidrule(lr){6-9} 
\textbf{Model} & ID & OOD & ID & OOD & TTFT ms $\downarrow$ & Tok./sec $\uparrow$ & VRAM (GB) $\downarrow$ & Input tok. $\downarrow$ \\ 
\midrule 
Prior sampling & 0.724 & \textcolor{gray}{0.622}$^{\dagger}$ & 0.635 & \textcolor{gray}{0.651}$^{\dagger}$ & -- & -- & -- & -- \\ 
Task-specific classifier & 0.752 & \textcolor{gray}{0.687}$^{\dagger}$ & 0.756 & \textcolor{gray}{0.741}$^{\dagger}$ & -- & -- & -- & -- \\ 
Contrastive alignment & 0.461 & 0.509 & 0.418 & 0.225 & -- & -- & -- & -- \\ 
LLM SFT ($h = 1$) & 0.794 & 0.704 & 0.850 & 0.536 & 164.91 & 57.08 & 9.02 & 348.50 \\ 
LLM SFT ($h = 5$) & 0.811 & \underline{0.750} & \textbf{0.893} & \underline{0.605} & 683.96 & 47.11 & 11.20 & 1370.12 \\ 
\rowcolor{cyan!12}
MINT ($h = 1$) & \underline{0.830} & \textbf{0.805} & 0.814 & \textbf{0.621} & \textbf{89.22} & \underline{138.83} & \textbf{7.66} & \textbf{140.50} \\ 
MINT ($h = 5$) & \textbf{0.832} & 0.696 & \underline{0.861} & 0.600 & \underline{134.69} & \textbf{136.80} & \underline{8.00} & \underline{258.12} \\
\bottomrule 
\end{tabular}
}
\end{table*}

\begin{table}[t]
\caption{Ablation study of MINT. We analyze the effects of input representation, training data composition, decoding strategy, transaction embedding design, and model capacity on extractive and predictive QA performance.}
\label{tab:qa_ablation_common}
\centering
\small
\resizebox{0.45\textwidth}{!}{
\begin{tabular}{lrrrr} 
\toprule 
& \multicolumn{2}{c}{\textbf{Predictive QA}} & \multicolumn{2}{c}{\textbf{Extractive QA}} \\ 
\cmidrule(lr){2-3}\cmidrule(lr){4-5} 
\textbf{Model} & ID & OOD & ID & OOD \\ 
\midrule 
\multicolumn{5}{l}{\textit{Transactions in the context prompt}} \\ 
with timestamp & \textbf{0.832} & 0.696 & \textbf{0.861} & 0.600 \\ 
without timestamp & 0.826 & \textbf{0.805} & 0.824 & \textbf{0.607} \\ 
\midrule 
\multicolumn{5}{l}{\textit{Decoding strategy}} \\ 
deterministic decoding & \textbf{0.832} & 0.696 & \textbf{0.861} & \textbf{0.600} \\ 
stochastic decoding & 0.814 & \textbf{0.700} & 0.858 & 0.595 \\ 
\midrule 
\multicolumn{5}{l}{\textit{Training data mixture}} \\ 
with CoT data & \textbf{0.832} & 0.696 & \textbf{0.861} & \textbf{0.600} \\ 
without CoT data & 0.826 & \textbf{0.722} & 0.742 & 0.439 \\ 
\midrule 
\multicolumn{5}{l}{\textit{Number of transaction embeddings (final layer)}} \\ 
$h = 1$ & 0.830 & \textbf{0.805} & 0.814 & \textbf{0.621} \\ 
$h = 5$ & \textbf{0.832} & 0.696 & 0.861 & 0.600 \\ 
$h = 10$ & \textbf{0.832} & 0.766 & 0.876 & 0.597 \\ 
$h = 25$ & 0.830 & 0.715 & \textbf{0.890} & 0.601 \\ 
\midrule 
\multicolumn{5}{l}{\textit{Number of transaction embeddings (second-to-last layer)}} \\ 
$h = 1$ & 0.834 & 0.688 & 0.855 & 0.553 \\ 
$h = 5$ & \textbf{0.839} & 0.659 & 0.897 & \textbf{0.609} \\ 
$h = 10$ & 0.836 & \textbf{0.698} & 0.907 & 0.591 \\ 
$h = 25$ & 0.838 & 0.696 & \textbf{0.916} & 0.578 \\ 
\midrule 
\multicolumn{5}{l}{\textit{LoRA rank}} \\ 
$r = 32$ & \textbf{0.834} & \textbf{0.749} & \textbf{0.862} & 0.583 \\ 
$r = 64$ & 0.832 & 0.696 & 0.861 & 0.600 \\ 
$r = 128$ & 0.825 & 0.657 & 0.858 & \textbf{0.609} \\ 
\midrule 
\multicolumn{5}{l}{\textit{Connector hidden layer size}} \\ 
$\ell = 512$ & 0.830 & \textbf{0.731} & \textbf{0.867} & \textbf{0.629} \\ 
$\ell = 1024$ & \textbf{0.832} & 0.696 & 0.861 & 0.600 \\ 
$\ell = 2048$ & 0.831 & 0.699 & 0.866 & 0.606 \\ 
\bottomrule 
\end{tabular}
}
\end{table}

\begin{table}[ht]
\caption{Impact of LLM-LTM pairing on MINT performance (overall accuracy). Results are reported for extractive and predictive QA under both ID and OOD evaluation settings across different Qwen3 backbone sizes and pretrained LTMs.}
\label{tab:qa_llm_ltm}
\centering
\small
\resizebox{0.37\textwidth}{!}{
\begin{tabular}{lrrrr} 
\toprule 
& \multicolumn{2}{c}{\textbf{Predictive QA}} & \multicolumn{2}{c}{\textbf{Extractive QA}} \\ 
\cmidrule(lr){2-3}\cmidrule(lr){4-5} 
\textbf{Model} & ID & OOD & ID & OOD \\ 
\midrule \multicolumn{5}{l}{\textit{Qwen3 0.6B}} \\ 
LTM 20M & 0.832 & \textbf{0.675} & 0.858 & 0.362 \\ 
LTM 120M & 0.831 & 0.632 & \textbf{0.861} & \textbf{0.592} \\ 
LTM 720M & \textbf{0.835} & 0.640 & 0.859 & 0.545 \\ 
\midrule \multicolumn{5}{l}{\textit{Qwen3 1.7B}} \\ 
LTM 20M & \textbf{0.832} & 0.696 & \textbf{0.861} & 0.600 \\ 
LTM 120M & 0.829 & \textbf{0.752} & 0.860 & \textbf{0.608} \\ 
LTM 720M & 0.825 & 0.727 & 0.853 & 0.592 \\ 
\midrule \multicolumn{5}{l}{\textit{Qwen3 4B}} \\ 
LTM 20M & 0.827 & \textbf{0.738} & \textbf{0.874} & 0.597 \\ 
LTM 120M & \textbf{0.830} & 0.729 & 0.873 & \textbf{0.614} \\ 
LTM 720M & 0.828 & 0.678 & 0.871 & 0.613 \\ 
\bottomrule 
\end{tabular}
}
\end{table}
%%%%%%%%%%%%%%%%%%%%%%%%%%%%%%%%%%%%% Table %%%%%%%%%%%%%%%%%%%%%%%%%%%%%%%%%%%%%

We evaluate extractive and predictive QA, each with in-distribution (ID) and out-of-distribution (OOD) splits, where OOD measures zero-shot generalization to unseen question types within the same transaction domain. Unless otherwise specified, MINT employs a Qwen3-1.7B instruction-tuned LLM \cite{qwen3}, a 20M-parameter transaction sequence encoder, and a two-layer MLP modality connector with hidden dimension 1024. The encoder produces 256-dimensional transaction embeddings, and the final $h=5$ transaction embeddings from the last encoder layer are projected into the 2048-dimensional LLM embedding space. To study scaling behavior, we additionally instantiate MINT with Qwen3 models at two other sizes (0.6B and 4B), enabling a controlled analysis of scaling across both the language backbone and the transaction sequence encoder.

For modality alignment, we use an equal mixture of structured report and semantic caption examples. During SFT, the LLM is adapted using LoRA applied to attention and MLP projections (rank \(r=64\), scaling \(\alpha=2 r\), dropout \(=0.05\)). The default SFT mixture consists of structured report (12.5\%), semantic caption (12.5\%), extractive QA (25\%), predictive QA (25\%), extractive QA with CoT rationales (12.5\%), and predictive QA with CoT rationales (12.5\%). In the \textit{w/o CoT} ablation, CoT examples are removed and the remaining tasks are reweighted proportionally.

Both modality alignment and SFT are optimized with AdamW using a learning rate of $10^{-4}$, $(\beta_1,\beta_2)=(0.9,0.95)$, weight decay $0.01$, warmup ratio $0.03$, gradient clipping of $1.0$, and mixed-precision bf16 training. We train on four A100 GPUs with an effective batch size of 128 samples per optimization step using gradient checkpointing throughout. Modality alignment is performed for two epochs on 160k samples per epoch, whereas SFT is run for a maximum of five epochs on 320k samples per epoch, with early stopping based on validation performance.

\textbf{Baselines.} Our primary baseline is \textbf{LLM SFT}, which serializes transaction histories as structured text records and is fine-tuned using the same LLM, LoRA rank, training data, and, where applicable, the training procedure, prompt template, and hyperparameter settings as MINT. Note that $h$ denotes the number of serialized transactions for LLM SFT and the number of injected transaction embeddings for MINT. For both LLM SFT and MINT, we use a maximum training sequence length of 2048 tokens per example. Due to this token budget, LLM SFT cannot accommodate transaction histories with $h \geq 10$. In contrast, MINT compresses transaction histories into a fixed-size representation, enabling efficient support for substantially longer histories. We also report \textbf{prior sampling} (sampling answers according to their empirical training frequencies), \textbf{task-specific classifiers} (classification heads trained on frozen transaction embeddings), and a \textbf{contrastive alignment} model that aligns transaction and caption embeddings using InfoNCE~\citep{fadeev2025latte,zhang2025sensorlm}. 

\vspace{-2mm}
\subsection{Main Results}
\label{subsec:results}

Table~\ref{tab:qa_main} reports QA accuracy together with inference metrics that contextualize the accuracy-efficiency trade-off.

\textbf{Predictive QA.} Under both ID and OOD evaluation, MINT ($h=1$) outperforms LLM SFT ($h=1$, $h=5$). % on both accuracy and macro-F1. 
This is the central empirical result of the paper: a single injected transaction embedding is sufficient to outperform text-serialized histories on the harder, more consequential task of forecasting future behavior.

\textbf{Extractive QA.} Under ID evaluation, LLM SFT ($h=1$, $h=5$) is relatively stronger than MINT ($h=1$), likely because the richer surface-level information in serialized text directly benefits extraction-style questions. Both MINT and LLM SFT exhibit monotonic performance gains with increasing history length $h$. Under OOD evaluation, MINT ($h=1$) surpasses LLM SFT ($h=1$, $h=5$).% in accuracy but falls short on macro-F1.

\textbf{Effect of history length.} Across all tasks, LLM SFT performance increases monotonically with $h$, at the cost of growing inference overhead. MINT does not exhibit this monotonic behavior except in extractive QA (ID); for predictive QA and OOD tasks, additional history embeddings provide no consistent benefit, and can even hurt performance, indicating that MINT extracts the relevant signal from a compact transaction representation.

\textbf{Additional baselines.} Both LLM SFT and MINT outperform prior sampling and task-specific classifiers on ID QA tasks. %, with one exception: the task-specific classifier attains a higher macro-F1 on predictive QA (ID). 
Prior sampling and task-specific classifiers are not directly applicable in the OOD setting, as they require training on the target questions; the OOD results reported for these methods are therefore included for reference only. The substantially lower accuracy of the contrastive alignment baseline indicates that alignment alone is insufficient for effective QA, highlighting the importance of task-specific supervision beyond caption training.

\textbf{Inference efficiency.} We measure time-to-first-token (TTFT), decode throughput (tokens/sec), and peak VRAM at a fixed batch size of 8 while varying history length $h$, comparing embedding injection (MINT) against tokenized serialization (LLM SFT). Although increasing $h$ increases computational cost for both approaches, MINT scales substantially more efficiently. Notably, MINT ($h=5$) remains more efficient than LLM SFT ($h=1$), achieving 18\% lower TTFT, 2.4$\times$ higher decode throughput, 11\% lower peak VRAM, and 26\% fewer input tokens. 

\vspace{-2mm}
\subsection{Ablations and Analysis}
\label{subsec:ablations}

We conduct ablations to understand the impact of architectural design, training strategy, and data composition (Tables~\ref{tab:qa_ablation_common} and \ref{tab:qa_llm_ltm}).

\textbf{Timestamps in the prompt.} Including the timestamp alongside each transaction embedding in the context prompt considerably improves extractive QA (ID) performance, %(accuracy and macro-F1), 
but considerably hurts predictive QA (OOD) performance. Extractive QA (OOD) and predictive QA (ID) are largely insensitive to this choice.

\textbf{Decoding strategy.} Deterministic (temperature $0$) and stochastic (temperature $0.7$, top-$p=0.9$) decoding yield roughly similar performance overall, with deterministic decoding slightly favored on the predictive QA (ID) task.

\textbf{Rationale supervision.} Including CoT rationales in the SFT mixture improves performance broadly, especially extractive QA, even though we do not prompt for explicit rationales at inference.

\textbf{History length and representation layer.} Increasing $h$ yields monotonic gains in extractive QA (ID) for both final- and second-to-last-layer embeddings, while predictive QA (ID) remains largely unchanged. In contrast, OOD performance exhibits no clear dependence on history length. Second-to-last layer embeddings consistently outperform final-layer embeddings on ID tasks but are generally weaker on OOD tasks, revealing a trade-off between ID accuracy and OOD robustness across representation depths.

\textbf{Capacity.} LoRA rank $r\in\{32,64,128\}$ and connector hidden size $\ell\in\{512,1024,2048\}$ have only modest effects on ID performance but larger effects on OOD. Among the configurations evaluated, $r=32$ and $\ell=512$ yield the strongest predictive OOD performance, indicating that increased capacity does not consistently improve generalization to unseen question types.

\textbf{Encoder/LLM capacity scaling (Table~\ref{tab:qa_llm_ltm}).} We jointly vary transaction sequence encoder size (20M / 120M / 720M) and LLM scale (0.6B-4B). On extractive QA (ID), the 4B LLM yields slightly better performance; on predictive QA (ID), all three LLM sizes perform similarly. For both ID tasks, performance is largely robust to transaction sequence encoder size at a given LLM size. On OOD tasks, the 0.6B LLM is noticeably weaker, while the 1.7B model performs best overall; for the 1.7B model specifically, the 120M transaction sequence encoder yields the best OOD performance. Taken together, Qwen3-1.7B offers the most favorable balance of quality and inference cost among the scales tested.

\textbf{General-purpose reasoning.} To assess retention of general-purpose reasoning, we evaluate the LoRA-adapted models on standard LLM benchmarks. Both MINT and LLM SFT underperform the base model, with ARC-Challenge accuracy dropping from 0.425 to 0.306/0.332, GSM8K from 0.412 to 0.026/0.033, and BoolQ from 0.776 to 0.493/0.492 (MINT/LLM SFT). This specialization cost arises only when LoRA adapters are enabled; disabling them immediately restores the base LLM's capabilities.

% !TEX root = main.tex
%%%%%%%%%%%%%%%%%%%%%%%%%%%%%%%%%%%%%
%%%%%%%%%%%%%%%%%%%%%%%%%%%%%%%%%%%%%
\vspace{-2mm}
\section{Conclusion}\label{sec:conclusion}
\label{sec:conclusion}

We presented \textsc{MINT}, a large-scale empirical study of how pretrained transaction embeddings interact with LLM-based reasoning. Across QA tasks, generalization, efficiency, and architectural choices, we find that transaction embeddings and text serialization occupy complementary regimes: the former excels in predictive QA and efficiency, while the latter remains stronger for extractive QA. We hypothesize that this predictive advantage arises from the encoder's autoregressive pretraining objective, which is naturally aligned with forecasting future behavior.

\textbf{Limitations.} Our experiments use proprietary transaction data that cannot be released due to privacy and regulatory constraints, although we provide detailed descriptions of the datasets, prompts, architectures, training procedures, and hyperparameters. More broadly, transaction reasoning systems introduce challenges related to privacy, profiling, fairness, and misuse, necessitating strong privacy safeguards, auditing, bias evaluation, access controls, and human oversight in real-world deployments.

\textbf{Future work.}
Several promising directions remain. First, we plan to extend beyond structured QA toward open-ended transaction reasoning and discovery tasks, including comparative analysis of customer behavior, context-aware behavioral assessment, and fraud-pattern discovery. Second, improving reliability through confidence-aware reasoning, e.g., generating \textit{(answer, rationale, confidence)} tuples, may lead to better calibrated decision-support systems. Third, reinforcement-based post-training offers a promising avenue for improving reasoning quality while maintaining compact multimodal representations. Finally, we intend to explore hybrid approaches that combine transaction embeddings with selectively retrieved textual context and prompt-compression techniques, potentially improving both reasoning performance and scalability.

%\vfill 
\vspace{2mm}
\begin{tcolorbox}[float,colback=gray!4,colframe=gray!40,title=\textbf{Summary of Key Findings}]
\small

\textbf{F1. Predictive reasoning benefits from transaction embeddings.}
MINT ($h=1$) outperforms LLM SFT on predictive QA, suggesting that transaction embeddings capture predictive signals more effectively than textual serialization.

\vspace{0.5em}

\textbf{F2. Text serialization is strong for extractive QA.}
Serialized transaction histories benefit extractive QA, especially as the number of transactions in the prompt increases.

\vspace{0.5em}

\textbf{F3. Embedding injection scales more efficiently than text serialization.}
MINT requires substantially fewer input tokens, lower latency, and less GPU memory.

\vspace{0.5em}

\textbf{F4. Additional transaction history is task dependent.}
Increasing history length consistently improves extractive QA (ID) but offers limited or negative gains for predictive QA.

\vspace{0.5em}

\textbf{F5. ID and OOD settings prefer different representations.}
Second-to-last encoder-layer embeddings consistently improve ID performance, whereas final-layer embeddings generally generalize better to unseen question types.

\vspace{0.5em}

%\textbf{F6. Chain-of-thought supervision improves performance.} Adding CoT data to the training mixture improves performance broadly, particularly for extractive QA.

\textbf{F6. Chain-of-thought supervision improves extractive QA.} Adding CoT data to the training mixture yields substantial gains on extractive QA, modest improvements on predictive QA (ID), and a slight drop on predictive QA (OOD).

\end{tcolorbox}

%\clearpage
\bibliographystyle{ACM-Reference-Format}
\bibliography{main}

\clearpage
\onecolumn
\appendix
% !TEX root = main.tex
%%%%%%%%%%%%%%%%%%%%%%%%%%%%%%%%%%%%%
%%%%%%%%%%%%%%%%%%%%%%%%%%%%%%%%%%%%%

\section{Datasets}
\label{subsec:data-construction-app}

\textbf{Prompt Templates.}
We represent a transaction history using special tokens interleaved with injected transaction embeddings. We add \texttt{<txn>}, \texttt{</txn>}, and \texttt{<emb>} to the tokenizer vocabulary. The \texttt{<emb>} token is a placeholder whose embedding is replaced at runtime by projected transaction vectors. We use a fixed-length history of $N$ transactions per example.
\begin{promptbox}
\small
\textcolor{gray!70}{Input:} \\
Transaction history of a customer: \\
Transaction at \{time\}: \texttt{<txn><emb></txn>} \\
Transaction at \{time\}: \texttt{<txn><emb></txn>} \\
\texttt{...}
\end{promptbox}
All tasks are expressed in a unified instruction-following format with structured output tags.

{Captioning:}
\begin{promptbox}
\small
\textcolor{gray!70}{Instruction:} \\
Write a structured report (or semantic description) of the transaction history. \\
Provide your summary in this format: \texttt{<summary>...</summary>} \\
\textcolor{gray!70}{Response:}
\end{promptbox}

{QA with rationale:}
\begin{promptbox}
\small
\textcolor{gray!70}{Instruction:} \\
Answer the following question based on the transaction history. \\
Provide your answer in this format: \texttt{<answer>...</answer>} \\
Provide your reasoning in this format: \texttt{<reason>...</reason>} \\
Question: \{question\} \\
\textcolor{gray!70}{Response:}
\end{promptbox}

\noindent\textbf{Extractive QA (ID):}
\begin{enumerate}
\item What is the value range of this transaction?
\item How does this transaction amount compare to the customer's average transaction amount?\item How does this transaction amount compare to the merchant's average transaction amount?\item Is this a card present or card not present transaction?
\item What is the merchant category of this transaction?
\item Has the customer transacted with this merchant before?   
\item How does this merchant category compare to this customer's most common spending category? 
\item What time of day did this transaction occur? 
\item On what type of day did this transaction occur?  
\item How much time elapsed between this transaction and the immediately preceding transaction?  
\item How many transactions has this customer made in the past {W}?      
\item Is the customer transacting at an unusually high velocity right now compared to their historical norm ({W} window)?         
\item How recently was the most recent declined transaction for this customer?         
\item How many declined transactions has this customer had in the past {W}?    
\item Does this transaction appear to be a retry of a recently declined transaction in the past {W} (same merchant, similar amount)? 
\item What is the customer's total spending level in the past {W}?        
\item What is the customer's average transaction amount over the past {W}?       
\item What has been the most frequent merchant category in the customer's transactions over the past {W}?            
\item How many distinct merchants has the customer visited in the past {W}? 
\end{enumerate}
where $W \in \{1h, 24h, 7d, 30d\}$.

\noindent\textbf{Extractive QA (OOD):}
\begin{enumerate}
\item In which country did this transaction take place?          
\item Has the customer made any large transactions (above $1,000$) in the past {W}?   
\item How many distinct countries has the customer transacted in over the past {W}? 
\end{enumerate}

\noindent\textbf{Predictive QA (ID):}
\begin{enumerate}
\item Is this transaction potentially fraudulent? 
\item Is this customer likely to be defrauded over the next {W}? 
\item What will be the merchant type of the next transaction? 
\item How much is this customer likely to spend over the next {W}? 
\item How many transactions is this customer likely to make in the next {W}? 
\item Is this customer likely to spend money on the same merchant category over the next {W}? 
\item Will the customer's spend in the same merchant category exceed $1000$ in the next {W}? 
\item What will be the most frequent merchant category for this customer over the next {W}? 
\end{enumerate}

\noindent\textbf{Predictive QA (OOD):}
\begin{enumerate}
\item What will be the country of the next transaction?  
\item Will the next transaction occur at the same merchant as the current one?  
\item What will be the approximate average transaction amount for this customer over the next {W}?                    
\item Is this customer likely to make a large transaction (above $1,000$) in the next {W}?
\end{enumerate}

\noindent\textbf{Reasoning data generation prompt:}
\begin{promptbox}
\small
\textcolor{gray!70}{Transaction Information:}\\
Transaction amount: ... \\
Merchant category code: ... \\
Merchant country: ... \\
Card present: ... \\
Transaction datetime: ... \\
Time since previous transaction (seconds): ... \\
Time since previous declined transaction (seconds): ... \\
Ratio of transaction amount to merchant mean: ... \\
Ratio of transaction amount to merchant max: ... \\

\textcolor{gray!70}{Customer History (past 30 days):}\\
Transactions in past 1h / 24h / 7d / 30d: ... / ... / ... / ... \\
Total spending in past 1h / 24h / 7d / 30d: ... / ... / ... / ... \\
Average transaction amount (30d): ... \\
Declined transactions in past 1h / 24h / 7d / 30d: ... / ... / ... / ... \\
Large transactions (>1000) in past 30d: ... \\
Distinct merchants (30d): ... \\
Distinct countries (30d): ... \\
Transactions at same merchant (30d): ... \\
Declined transactions at same merchant with similar amount (1h / 24h): ... / ... \\

\textcolor{gray!70}{Question and Answer:}\\
Question: ... \\
Correct Answer: ... \\

Given the transaction information and customer history above, provide a clear step-by-step explanation of why the given answer is correct. Consider the transaction properties, spending patterns, merchant history, and any relevant risk signals. \\

Use exactly this format: \\
Reason: \texttt{<your step-by-step reasoning>}
\end{promptbox}

\noindent\textbf{Semantic description generation prompt:}
\begin{promptbox}
\small
\textcolor{gray!70}{Task:}\\
Write a concise description of a cardholder's transaction behaviour based on a window of up to 3 recent transactions, commenting on any unusual or fraudulent behaviour observed. \\

\textcolor{gray!70}{Transaction 1:}\\
Entityid: ... \\
Transaction amount: ... \\
Time since previous transaction (seconds): ... \\
Time since previous declined transaction (seconds): ... \\
Card present: ... \\
Merchant category code: ... \\
Merchant country: ... \\
The ratio of the current transaction amount to the 90-day exponentially smoothed mean transaction amount at the merchant: ... \\
The ratio of the current transaction amount to the 90-day exponentially smoothed max transaction amount at the merchant: ... \\
The sum of transaction amounts by the cardholder exponentially decayed over the previous 1 hour: ... \\
The sum of transaction amounts by the cardholder exponentially decayed over the previous 1 day: ... \\

\textcolor{gray!70}{Transaction 2:}\\
... \\

\textcolor{gray!70}{Transaction 3:}\\
... \\

\textcolor{gray!70}{Description:}\\
\end{promptbox}

\section{Training}
\label{sec:training-app}

Let $x=(\mathbf{T},\mathbf{q},\mathbf{y})$ denote a training example, where $\mathbf{T}$ is a transaction history, $\mathbf{q}$ is an instruction or question, and $\mathbf{y}=(y_1,\dots,y_{|\mathbf{y}|})$ is the target response. The frozen encoder produces $e_{1:N} = E(\mathbf{T})$, which is mapped by the connector $g_\phi$ to injected embeddings $z_{1:N} = g_\phi(e_{1:N})$. Conditioning on $\mathbf{q}$ and the injected embeddings, the decoder-only LLM models
\begin{equation}
p_{\theta,\Delta}(\mathbf{y} \mid \mathbf{T}, \mathbf{q}) 
~=~ \prod_{t=1}^{|\mathbf{y}|} p_{\theta,\Delta}(y_t \mid \mathbf{y}_{<t}, z_{1:N}, \mathbf{q}),
\end{equation}
where $\theta$ are frozen base parameters and $\Delta$ are LoRA parameters (when enabled).

\textbf{Modality Alignment.}
The goal is to learn $g_\phi$ such that injected transaction embeddings are compatible with the LLM token representations. We train the connector on captioning data $\mathcal{D}_{\text{cap}}$ using a completion-only autoregressive loss:
\begin{equation}
\mathcal{L}_{\text{align}}(\phi) 
~=~ 
\Expectover{(\mathbf{T}, \mathbf{q}, \mathbf{y}) \sim \mathcal{D}_{\text{cap}}}{- \sum_{t \in \mathcal{M}} \log  p_{\theta,\Delta=0}(y_t \mid \mathbf{y}_{<t}, z_{1:N}, \mathbf{q})} ,
\end{equation}
where $\mathcal{M}$ denotes the index set corresponding to response tokens (excluding prompt tokens). During alignment, $\theta$ is frozen and LoRA is disabled ($\Delta = 0$); only $\phi$ is updated.

\textbf{Supervised Fine-Tuning (SFT).}
We then jointly optimize connector $\phi$ and LoRA $\Delta$ on a mixture of captioning and QA datasets, $\mathcal{D} = \mathcal{D}_{\text{cap}} \cup \mathcal{D}_{\text{qa}}$, using a completion-only autoregressive loss:
\begin{equation}
\mathcal{L}_{\text{sft}}(\phi, \Delta) 
~=~ 
\Expectover{(\mathbf{T}, \mathbf{q}, \mathbf{y}) \sim \mathcal{D}}{- \sum_{t \in \mathcal{M}} \log p_{\theta,\Delta}(y_t \mid \mathbf{y}_{<t}, z_{1:N}, \mathbf{q})} .
\end{equation}
We interleave captioning and QA to preserve generation quality while improving grounded reasoning. For QA examples with rationales, $\mathbf{y}$ concatenates the reasoning trace and final answer; for answer-only examples, $\mathbf{y}$ contains only the short answer.

\section{Additional Results}

Tables~\ref{tab:qa_main_detailed}, \ref{tab:qa_ablation_common_detailed}, and \ref{tab:qa_llm_ltm_detailed} provide detailed counterparts to Tables~\ref{tab:qa_main}, \ref{tab:qa_ablation_common}, and \ref{tab:qa_llm_ltm}, respectively. In addition to mean accuracy, they report macro-F1 and standard deviations computed across per-question-type average accuracy and macro-F1 scores, providing a more complete view of performance variability across question types. Notably, the contrastive alignment model achieves relatively strong macro-F1 despite substantially lower accuracy. This suggests that it treats answer classes more uniformly but lacks the task-specific supervision needed to achieve high overall correctness. Since the model is trained solely on caption-alignment data and never observes the imbalanced QA training set used by MINT and LLM-SFT, it is largely unaffected by QA label imbalance and exhibits little distinction between ID and OOD evaluation, effectively operating in a zero-shot QA setting.

\textbf{Training data balancing.} 
To study the impact of label imbalance, we construct a balanced QA training set by rescaling answer frequencies within each question type. Specifically, answer counts are normalized relative to the majority class, capped at 10,000 examples, and minority classes are guaranteed at least 2,000 examples whenever sufficient data are available. This strategy reduces extreme class imbalance while preserving the overall diversity of questions and answers. As shown in Table~\ref{tab:qa_ablation_common}, balancing improves macro-F1 on ID QA tasks with little or no loss in accuracy, indicating better performance on minority answer classes. However, it reduces both accuracy and macro-F1 on extractive QA (OOD), suggesting that the original answer distribution may provide useful inductive bias for generalization to unseen question types. For predictive QA (OOD), balancing increases macro-F1 at a modest cost to accuracy, reflecting the typical trade-off between overall correctness and more uniform performance across answer classes.

\textbf{Confidence and calibration analysis.} Table~\ref{tab:confidence} compares the confidence and calibration properties of MINT and LLM-SFT on predictive QA. Beyond accuracy, we evaluate the quality of the predicted probability distribution using several complementary metrics: mean confidence, entropy, and margin measure prediction sharpness; confidence on correct vs.\ incorrect predictions and the resulting confidence gap quantify how well confidence distinguishes correct from incorrect answers; ECE, MCE, and Brier score assess probability calibration; and AUROC measures the ability of confidence scores to separate correct and incorrect predictions. Across most question types, MINT achieves higher accuracy, larger margins, lower entropy, larger confidence gaps, lower calibration error, and higher AUROC than LLM-SFT. These results indicate that MINT not only produces more accurate predictions but also generates sharper, better-calibrated confidence estimates that more reliably reflect prediction correctness.

\textbf{Rationale-augmented prompting.} We further study prompting the model to produce answers only versus answers with explicit rationales (Table~\ref{tab:qa_answer_reason_prompt}). While including CoT data in the training mixture improves overall QA performance, explicitly prompting for <reason> and <answer> outputs at inference time does not translate into better downstream accuracy, and increases formatting failures. We attribute this to two factors: our CoT training data is small and its reasoning traces are long, so even with generous decoding budgets (1024-2048 max new tokens), the model frequently fails to emit a well-formed answer within budget; and fine-tuning on concise CoT examples alone does not reliably induce concise reasoning at inference time. This suggests CoT is most valuable as an implicit training signal, with explicit rationale generation likely requiring additional reinforcement learning (e.g., an accuracy plus length-penalty reward) to be practically usable.

\textbf{Caption and reasoning quality (Tables~\ref{tab:qa_main_cot}~and~\ref{tab:caption_main}).} We additionally evaluate caption and rationale quality using an LLM-as-judge. For transaction captioning, the judge scores outputs on faithfulness, coverage, specificity, coherence, and conciseness; for rationales, it evaluates clarity, depth, faithfulness, coherence, and relevance. The judge receives (i) a text-rendered transaction history, (ii) the instruction or question, (iii) the model output, and (iv) a reference answer when available, and returns a structured JSON evaluation. We note that this setup inherently favors the LLM SFT baseline, whose fully text-serialized transaction representation more closely matches the judge's input format. Despite this bias, MINT achieves comparable judge scores, indicating that embedding-based transaction representations preserve information quality while enabling more efficient modeling.

\noindent\textbf{Caption judge prompt:}
\begin{promptbox}
\small
\textbf{System:} You are an expert evaluator of transaction summaries. \\
\textbf{User:} You will be given (1) transaction history, (2) a candidate caption, and optionally (3) a reference caption. \\
\textbf{Task:} Score the candidate on: Faithfulness, Coverage, Specificity, Coherence, Conciseness (1--5). \\
\textbf{Rules:} \\
(1) The caption must be grounded in the transaction history. Penalize hallucinations heavily. \\
(2) If a reference is provided, use it only to assess missing key points; do not penalize alternative correct phrasings. \\
(3) Return valid JSON: \\
\{\texttt{"faithfulness":int, "coverage":int, "specificity":int, "coherence":int, "conciseness":int, "notes":string}\}. \\
\textbf{Input:} \\
\{transaction history\} \\
\{reference caption (optional)\} \\
\{candidate caption\}
\end{promptbox}

\noindent\textbf{Rationale judge prompt:}
\begin{promptbox}
\small
\textbf{System:} You are an expert evaluator of reasoning grounded in transaction histories. \\
\textbf{User:} You will be given (1) transaction history, (2) question, (3) true answer, and (4) candidate answer and rationale. \\
\textbf{Task:} Score the rationale on: Clarity, Depth, Faithfulness, Coherence, Relevance (1--5). \\
\textbf{Rules:} \\
(1) The rationale must cite or implicitly rely on facts present in the history; penalize unsupported claims. \\
(2) Prefer reasoning that uses correct temporal aggregation and attribute matching. \\
(3) Return valid JSON: \\
\{\texttt{"clarity":int, "depth":int, "faithfulness":int, "coherence":int, "relevance":int, "notes":string}\}. \\
\textbf{Input:} \\
\{transaction history\} \\
\{question\} \\
\{True answer\} \\
\{candidate answer\} \\
\{candidate rationale\}
\end{promptbox}

\begin{table*}[ht]
\centering
\small
\resizebox{\textwidth}{!}{
\begin{tabular}{lcccccccc}
\toprule
& \multicolumn{2}{c}{\textbf{Extractive QA (ID)}} & \multicolumn{2}{c}{\textbf{Predictive QA (ID)}} & \multicolumn{2}{c}{\textbf{Extractive QA (OOD)}} & \multicolumn{2}{c}{\textbf{Predictive QA (OOD)}} \\
\cmidrule(lr){2-3}\cmidrule(lr){4-5}\cmidrule(lr){6-7}\cmidrule(lr){8-9}
\textbf{Model} & Acc $\uparrow$ & Macro-F1 $\uparrow$ & Acc $\uparrow$ & Macro-F1 $\uparrow$ & Acc $\uparrow$ & Macro-F1 $\uparrow$ & Acc $\uparrow$ & Macro-F1 $\uparrow$ \\
\midrule
Prior sampling & 0.635 ± 0.250 & 0.301 ± 0.135 & 0.724 ± 0.251 & 0.374 ± 0.138 & \textcolor{gray}{0.651 ± 0.047}$^{\dagger}$ & \textcolor{gray}{0.332 ± 0.001}$^{\dagger}$ & \textcolor{gray}{0.622 ± 0.125}$^{\dagger}$ & \textcolor{gray}{0.329 ± 0.122}$^{\dagger}$ \\
Task-specific classifier & 0.756 ± 0.177 & 0.514 ± 0.184 & 0.752 ± 0.227 & \textbf{0.483 ± 0.155} & \textcolor{gray}{0.741 ± 0.012}$^{\dagger}$ & \textcolor{gray}{0.422 ± 0.042}$^{\dagger}$ & \textcolor{gray}{0.687 ± 0.117}$^{\dagger}$ & \textcolor{gray}{0.434 ± 0.106}$^{\dagger}$ \\
\textcolor{gray}{Contrastive alignment} & \textcolor{gray}{0.418 ± 0.180} & \textcolor{gray}{0.483 ± 0.214} & \textcolor{gray}{0.461 ± 0.196} & \textcolor{gray}{0.527 ± 0.208} & \textcolor{gray}{0.225 ± 0.200} & \textcolor{gray}{0.376 ± 0.376} & \textcolor{gray}{0.509 ± 0.131} & \textcolor{gray}{0.614 ± 0.167} \\
LLM SFT ($h = 1$) & 0.850 ± 0.172 & 0.532 ± 0.267 & 0.794 ± 0.219 & 0.409 ± 0.160 & 0.536 ± 0.386 & 0.281 ± 0.178 & 0.704 ± 0.281 & 0.381 ± 0.170 \\
LLM SFT ($h = 5$) & \textbf{0.893 ± 0.134} & \textbf{0.631 ± 0.253} & 0.811 ± 0.199 & 0.416 ± 0.154 & 0.605 ± 0.402 & \textbf{0.342 ± 0.214} & 0.750 ± 0.197 & 0.443 ± 0.173 \\
MINT ($h = 1$) & 0.814 ± 0.158 & 0.480 ± 0.206 & \textbf{0.830 ± 0.177} & 0.450 ± 0.155 & \textbf{0.621 ± 0.399} & 0.292 ± 0.167 & \textbf{0.805 ± 0.140} & \textbf{0.464 ± 0.144} \\
\bottomrule
\end{tabular}
}
\caption{Overall accuracy and macro-F1 on extractive and predictive QA tasks. Results marked with $^{\dagger}$ are trained and evaluated on the same question set and are included for reference only.}
\label{tab:qa_main_detailed}
\end{table*}

\begin{table*}[ht]
\centering
\small
\resizebox{\textwidth}{!}{
\begin{tabular}{lcccccccc}
\toprule
& \multicolumn{2}{c}{\textbf{Extractive QA (ID)}} & \multicolumn{2}{c}{\textbf{Predictive QA (ID)}} & \multicolumn{2}{c}{\textbf{Extractive QA (OOD)}} & \multicolumn{2}{c}{\textbf{Predictive QA (OOD)}} \\
\cmidrule(lr){2-3}\cmidrule(lr){4-5}\cmidrule(lr){6-7}\cmidrule(lr){8-9}
\textbf{Model} & Acc $\uparrow$ & Macro-F1 $\uparrow$ & Acc $\uparrow$ & Macro-F1 $\uparrow$ & Acc $\uparrow$ & Macro-F1 $\uparrow$ & Acc $\uparrow$ & Macro-F1 $\uparrow$ \\
\midrule
\multicolumn{9}{l}{\textit{Transactions in the context prompt}} \\
with timestamp & \textbf{0.861 ± 0.133} & \textbf{0.565 ± 0.188} & \textbf{0.832 ± 0.172} & \textbf{0.449 ± 0.147} & 0.600 ± 0.396 & \textbf{0.291 ± 0.173} & 0.696 ± 0.264 & 0.385 ± 0.141 \\
without timestamp & 0.824 ± 0.143 & 0.509 ± 0.164 & 0.826 ± 0.182 & 0.444 ± 0.159 & \textbf{0.607 ± 0.394} & 0.285 ± 0.168 & \textbf{0.805 ± 0.147} & \textbf{0.486 ± 0.177} \\
\midrule
\multicolumn{9}{l}{\textit{Decoding strategy}} \\
deterministic decoding & \textbf{0.861 ± 0.133} & 0.565 ± 0.188 & \textbf{0.832 ± 0.172} & \textbf{0.449 ± 0.147} & \textbf{0.600 ± 0.396} & 0.291 ± 0.173 & 0.696 ± 0.264 & 0.385 ± 0.141 \\
stochastic decoding & 0.858 ± 0.139 & \textbf{0.568 ± 0.193} & 0.814 ± 0.191 & 0.447 ± 0.143 & 0.595 ± 0.387 & \textbf{0.304 ± 0.176} & \textbf{0.700 ± 0.237} & \textbf{0.405 ± 0.126} \\
\midrule
\multicolumn{9}{l}{\textit{Training data mixture}} \\
with CoT data & \textbf{0.861 ± 0.133} & \textbf{0.565 ± 0.188} & \textbf{0.832 ± 0.172} & \textbf{0.449 ± 0.147} & \textbf{0.600 ± 0.396} & \textbf{0.291 ± 0.173} & 0.696 ± 0.264 & \textbf{0.385 ± 0.141} \\
without CoT data & 0.742 ± 0.194 & 0.371 ± 0.147 & 0.826 ± 0.178 & 0.415 ± 0.130 & 0.439 ± 0.304 & 0.230 ± 0.121 & \textbf{0.722 ± 0.233} & 0.383 ± 0.132 \\
% \midrule
% \multicolumn{9}{l}{\textit{Training data balancing ($h = 1$)}} \\
% with original QA / CoT data & \textbf{0.814 ± 0.158} & 0.480 ± 0.206 & \textbf{0.830 ± 0.177} & 0.450 ± 0.155 & \textbf{0.621 ± 0.399} & \textbf{0.292 ± 0.167} & \textbf{0.805 ± 0.140} & \textbf{0.464 ± 0.144} \\
% with balanced QA / CoT data & 0.803 ± 0.155 & \textbf{0.500 ± 0.188} & 0.825 ± 0.171 & \textbf{0.495 ± 0.148} & 0.599 ± 0.385 & 0.288 ± 0.155 & 0.703 ± 0.226 & 0.445 ± 0.115 \\
\midrule
\multicolumn{9}{l}{\textit{Training data balancing}} \\
with original QA / CoT data & 0.861 ± 0.133 & 0.565 ± 0.188 & \textbf{0.832 ± 0.172} & 0.449 ± 0.147 & \textbf{0.600 ± 0.396} & \textbf{0.291 ± 0.173} & \textbf{0.696 ± 0.264} & 0.385 ± 0.141 \\
with balanced QA / CoT data & \textbf{0.863 ± 0.128} & \textbf{0.601 ± 0.191} & 0.831 ± 0.161 & \textbf{0.496 ± 0.145} & 0.458 ± 0.362 & 0.234 ± 0.129 & 0.678 ± 0.247 & \textbf{0.434 ± 0.128} \\
\midrule
\multicolumn{9}{l}{\textit{Number of transaction embeddings (final layer)}} \\
$h = 1$ & 0.814 ± 0.158 & 0.480 ± 0.206 & 0.830 ± 0.177 & 0.450 ± 0.155 & \textbf{0.621 ± 0.399} & 0.292 ± 0.167 & \textbf{0.805 ± 0.140} & \textbf{0.464 ± 0.144} \\
$h = 5$ & 0.861 ± 0.133 & 0.565 ± 0.188 & \textbf{0.832 ± 0.172} & 0.449 ± 0.147 & 0.600 ± 0.396 & 0.291 ± 0.173 & 0.696 ± 0.264 & 0.385 ± 0.141 \\
$h = 10$ & 0.876 ± 0.125 & 0.601 ± 0.203 & 0.832 ± 0.174 & 0.450 ± 0.156 & 0.597 ± 0.401 & 0.291 ± 0.181 & 0.766 ± 0.190 & 0.427 ± 0.107 \\
$h = 25$ & \textbf{0.890 ± 0.111} & \textbf{0.606 ± 0.206} & 0.830 ± 0.177 & \textbf{0.452 ± 0.155} & 0.601 ± 0.397 & \textbf{0.293 ± 0.178} & 0.715 ± 0.228 & 0.415 ± 0.134 \\
%$h = 50$ & 0.887 ± 0.117 & 0.601 ± 0.205 & 0.838 ± 0.166 & 0.454 ± 0.148 & 0.608 ± 0.369 & 0.301 ± 0.159 & 0.756 ± 0.173 & 0.461 ± 0.133 \\
\midrule
\multicolumn{9}{l}{\textit{Number of transaction embeddings (second last layer)}} \\
$h = 1$ & 0.855 ± 0.144 & 0.537 ± 0.234 & 0.834 ± 0.180 & 0.458 ± 0.168 & 0.553 ± 0.377 & 0.265 ± 0.157 & 0.688 ± 0.255 & 0.405 ± 0.134 \\
$h = 5$ & 0.897 ± 0.110 & 0.618 ± 0.215 & \textbf{0.839 ± 0.167} & \textbf{0.470 ± 0.160} & \textbf{0.609 ± 0.402} & 0.279 ± 0.172 & 0.659 ± 0.299 & 0.386 ± 0.147 \\
$h = 10$ & 0.907 ± 0.103 & 0.656 ± 0.218 & 0.836 ± 0.169 & 0.467 ± 0.160 & 0.591 ± 0.387 & \textbf{0.298 ± 0.173} & \textbf{0.698 ± 0.229} & \textbf{0.436 ± 0.152} \\
$h = 25$ & \textbf{0.916 ± 0.092} & \textbf{0.658 ± 0.220} & 0.838 ± 0.170 & 0.466 ± 0.168 & 0.578 ± 0.385 & 0.286 ± 0.184 & 0.696 ± 0.226 & 0.416 ± 0.125 \\
\midrule
\multicolumn{9}{l}{\textit{LoRA rank}} \\
$r = 32$ & \textbf{0.862 ± 0.134} & 0.562 ± 0.189 & \textbf{0.834 ± 0.173} & \textbf{0.454 ± 0.153} & 0.583 ± 0.381 & 0.288 ± 0.154 & \textbf{0.749 ± 0.211} & \textbf{0.416 ± 0.113} \\
$r = 64$ & 0.861 ± 0.133 & \textbf{0.565 ± 0.188} & 0.832 ± 0.172 & 0.449 ± 0.147 & 0.600 ± 0.396 & \textbf{0.291 ± 0.173} & 0.696 ± 0.264 & 0.385 ± 0.141 \\
$r = 128$ & 0.858 ± 0.137 & 0.545 ± 0.187 & 0.825 ± 0.182 & 0.444 ± 0.156 & \textbf{0.609 ± 0.398} & 0.270 ± 0.160 & 0.657 ± 0.302 & 0.376 ± 0.171 \\
\midrule
\multicolumn{9}{l}{\textit{Connector hidden layer size}} \\
$\ell = 512$ & \textbf{0.867 ± 0.133} & \textbf{0.576 ± 0.199} & 0.830 ± 0.178 & 0.449 ± 0.145 & \textbf{0.629 ± 0.381} & \textbf{0.321 ± 0.171} & \textbf{0.731 ± 0.212} & \textbf{0.423 ± 0.136} \\
$\ell = 1024$ & 0.861 ± 0.133 & 0.565 ± 0.188 & \textbf{0.832 ± 0.172} & 0.449 ± 0.147 & 0.600 ± 0.396 & 0.291 ± 0.173 & 0.696 ± 0.264 & 0.385 ± 0.141 \\
$\ell = 2048$ & 0.866 ± 0.133 & \textbf{0.576 ± 0.196} & 0.831 ± 0.176 & 0.449 ± 0.150 & 0.606 ± 0.403 & 0.286 ± 0.179 & 0.699 ± 0.254 & 0.401 ± 0.145 \\
\bottomrule
\end{tabular}
}
\caption{Ablation study of MINT. We analyze the effects of input representation, training data composition, decoding strategy, transaction embedding design, and model capacity on extractive and predictive QA performance (overall accuracy and macro-F1) in both ID and OOD settings.}
\label{tab:qa_ablation_common_detailed}
\end{table*}

\begin{table*}[ht]
\centering
\small
\resizebox{0.9\textwidth}{!}{
\begin{tabular}{lcccccccc}
\toprule
& \multicolumn{2}{c}{\textbf{Extractive QA}} & \multicolumn{2}{c}{\textbf{Predictive QA}} & \multicolumn{2}{c}{\textbf{Extractive QA (OOD)}} & \multicolumn{2}{c}{\textbf{Predictive QA (OOD)}} \\
\cmidrule(lr){2-3}\cmidrule(lr){4-5}\cmidrule(lr){6-7}\cmidrule(lr){8-9}
\textbf{Model} & Acc $\uparrow$ & Macro-F1 $\uparrow$ & Acc $\uparrow$ & Macro-F1 $\uparrow$ & Acc $\uparrow$ & Macro-F1 $\uparrow$ & Acc $\uparrow$ & Macro-F1 $\uparrow$ \\
\midrule
\multicolumn{9}{l}{\textit{Qwen3 0.6B}} \\
LTM 20M & 0.858 ± 0.138 & 0.562 ± 0.193 & 0.832 ± 0.174 & 0.456 ± 0.150 & 0.362 ± 0.393 & 0.161 ± 0.136 & \textbf{0.675 ± 0.280} & \textbf{0.389 ± 0.170} \\
LTM 120M & \textbf{0.861 ± 0.137} & \textbf{0.567 ± 0.194} & 0.831 ± 0.177 & 0.447 ± 0.158 & \textbf{0.592 ± 0.394} & 0.270 ± 0.170 & 0.632 ± 0.345 & 0.348 ± 0.181 \\
LTM 720M & 0.859 ± 0.135 & 0.565 ± 0.188 & \textbf{0.835 ± 0.166} & \textbf{0.456 ± 0.150} & 0.545 ± 0.368 & \textbf{0.280 ± 0.178} & 0.640 ± 0.289 & 0.383 ± 0.168 \\
\midrule
\multicolumn{9}{l}{\textit{Qwen3 1.7B}} \\
LTM 20M & \textbf{0.861 ± 0.133} & 0.565 ± 0.188 & \textbf{0.832 ± 0.172} & 0.449 ± 0.147 & 0.600 ± 0.396 & 0.291 ± 0.173 & 0.696 ± 0.264 & 0.385 ± 0.141 \\
%LTM 20M & 0.863 ± 0.130 & 0.568 ± 0.188 & 0.833 ± 0.169 & 0.457 ± 0.147 & 0.416 ± 0.445 & 0.185 ± 0.157 & 0.697 ± 0.279 & 0.384 ± 0.162 \\
LTM 120M & 0.860 ± 0.136 & \textbf{0.576 ± 0.194} & 0.829 ± 0.180 & 0.450 ± 0.152 & \textbf{0.608 ± 0.404} & \textbf{0.309 ± 0.196} & \textbf{0.752 ± 0.181} & \textbf{0.449 ± 0.123} \\
LTM 720M & 0.853 ± 0.141 & 0.559 ± 0.188 & 0.825 ± 0.178 & \textbf{0.452 ± 0.154} & 0.592 ± 0.384 & 0.293 ± 0.174 & 0.727 ± 0.218 & 0.414 ± 0.127 \\
\midrule
\multicolumn{9}{l}{\textit{Qwen3 4B}} \\
LTM 20M & \textbf{0.874 ± 0.129} & 0.601 ± 0.209 & 0.827 ± 0.178 & \textbf{0.458 ± 0.156} & 0.597 ± 0.404 & \textbf{0.323 ± 0.216} & \textbf{0.738 ± 0.242} & \textbf{0.412 ± 0.151} \\
LTM 120M & 0.873 ± 0.129 & 0.590 ± 0.203 & \textbf{0.830 ± 0.175} & 0.453 ± 0.154 & \textbf{0.614 ± 0.391} & 0.308 ± 0.183 & 0.729 ± 0.260 & 0.395 ± 0.143 \\
LTM 720M & 0.871 ± 0.131 & \textbf{0.603 ± 0.198} & 0.828 ± 0.175 & 0.450 ± 0.148 & 0.613 ± 0.396 & 0.295 ± 0.170 & 0.678 ± 0.302 & 0.372 ± 0.180 \\
\bottomrule
\end{tabular}
}
\caption{Impact of LLM-LTM pairing on MINT performance (overall accuracy and macro-F1). Results are reported for extractive and predictive QA under both ID and OOD evaluation settings across different Qwen3 backbone sizes and pretrained LTMs.}
\label{tab:qa_llm_ltm_detailed}
\end{table*}

\begin{table*}[t]
\centering
\small
\resizebox{\textwidth}{!}{
\begin{tabular}{lccccccccccccc}
\toprule
Metric & (6) 1h & (6) 24h & (6) 30d & (6) 7d & (8) 1h & (8) 24h & (8) 30d & (8) 7d & (3) & (4) 7d & (5) 24h & (5) 30d & (5) 7d \\
\midrule
Accuracy $\uparrow$ 
& \textbf{0.903}/0.898 & \textbf{0.729}/0.713 & \textbf{0.857}/0.846 & \textbf{0.753}/0.728 
& \textbf{0.865}/0.864 & \textbf{0.554}/0.551 & \textbf{0.507}/0.444 & \textbf{0.462}/0.398 
& \textbf{0.528}/0.517 & \textbf{0.568}/0.460 & \textbf{0.710}/0.652 & \textbf{0.787}/0.759 & \textbf{0.699}/0.621 \\

Mean Conf $\uparrow$ 
& 0.893/\underline{0.915} & 0.739/\underline{0.746} & \textbf{0.867}/0.842 & \textbf{0.776}/0.734 
& \textbf{0.862}/0.857 & 0.555/\underline{0.581} & \textbf{0.756}/0.653 & \textbf{0.627}/0.577 
& 0.508/\underline{0.525} & \textbf{0.688}/0.584 & \textbf{0.740}/0.702 & \textbf{0.883}/0.789 & \textbf{0.780}/0.723 \\

Entropy $\downarrow$ 
& 0.437/\underline{0.386} & 0.755/\underline{0.751} & \textbf{0.417}/0.509 & \textbf{0.683}/0.772 
& \textbf{0.258}/0.278 & 0.557/\underline{0.542} & \textbf{0.314}/0.421 & \textbf{0.473}/0.502 
& 0.602/0.602 & \textbf{0.642}/0.807 & \textbf{0.393}/0.433 & \textbf{0.202}/0.364 & \textbf{0.379}/0.443 \\

Margin $\uparrow$ 
& 0.786/\underline{0.831} & 0.478/\underline{0.492} & \textbf{0.735}/0.685 & \textbf{0.553}/0.468 
& \textbf{0.783}/0.780 & 0.314/\underline{0.350} & \textbf{0.572}/0.420 & \textbf{0.394}/0.316 
& 0.255/\underline{0.280} & \textbf{0.471}/0.313 & \textbf{0.486}/0.411 & \textbf{0.781}/0.602 & \textbf{0.591}/0.469 \\

Conf (correct) $\uparrow$ 
& 0.901/\underline{0.921} & 0.763/\underline{0.766} & \textbf{0.895}/0.869 & \textbf{0.800}/0.753 
& \textbf{0.869}/0.860 & 0.589/\underline{0.602} & \textbf{0.778}/0.686 & \textbf{0.650}/0.601 
& 0.540/\underline{0.544} & \textbf{0.737}/0.616 & \textbf{0.767}/0.722 & \textbf{0.917}/0.820 & \textbf{0.798}/0.736 \\

Conf (incorrect) $\downarrow$ 
& \textbf{0.813}/0.867 & \textbf{0.673}/0.696 & 0.698/0.697 & 0.704/\underline{0.683} 
& \textbf{0.816}/0.833 & \textbf{0.513}/0.555 & 0.732/\underline{0.627} & 0.608/\underline{0.561} 
& \textbf{0.473}/0.504 & 0.624/\underline{0.557} & 0.675/\underline{0.665} & 0.760/\underline{0.691} & 0.739/\underline{0.700} \\

Conf Gap $\uparrow$ 
& \textbf{0.089}/0.053 & \textbf{0.090}/0.070 & \textbf{0.197}/0.172 & \textbf{0.096}/0.070 
& \textbf{0.053}/0.028 & \textbf{0.077}/0.047 & 0.046/\underline{0.059} & \textbf{0.042}/0.040 
& \textbf{0.067}/0.040 & \textbf{0.113}/0.058 & \textbf{0.092}/0.057 & \textbf{0.157}/0.130 & \textbf{0.060}/0.036 \\

ECE $\downarrow$ 
& \textbf{0.010}/0.022 & \textbf{0.011}/0.033 & \textbf{0.010}/0.012 & 0.023/\underline{0.014} 
& \textbf{0.015}/0.016 & \textbf{0.019}/0.043 & 0.249/\underline{0.209} & \textbf{0.165}/0.179 
& 0.029/\underline{0.022} & \textbf{0.121}/0.124 & \textbf{0.030}/0.051 & 0.097/\underline{0.030} & \textbf{0.081}/0.102 \\

MCE $\downarrow$ 
& \textbf{0.037}/0.500 & \textbf{0.072}/0.500 & 0.500/\underline{0.056} & 0.300/\underline{0.167} 
& 0.721/\underline{0.262} & \textbf{0.088}/0.091 & 0.359/\underline{0.319} & 0.381/\underline{0.297} 
& 0.069/\underline{0.049} & \textbf{0.191}/0.213 & \textbf{0.058}/0.066 & \textbf{0.277}/0.398 & \textbf{0.197}/0.238 \\

Brier $\downarrow$ 
& \textbf{0.081}/0.087 & \textbf{0.180}/0.193 & \textbf{0.098}/0.107 & \textbf{0.170}/0.186 
& \textbf{0.034}/0.035 & \textbf{0.083}/0.087 & \textbf{0.102}/0.105 & \textbf{0.100}/0.106 
& \textbf{0.086}/0.090 & \textbf{0.190}/0.221 & \textbf{0.095}/0.110 & \textbf{0.078}/0.085 & \textbf{0.111}/0.132 \\

AUROC $\uparrow$ 
& \textbf{0.722}/0.640 & \textbf{0.693}/0.659 & \textbf{0.840}/0.812 & \textbf{0.697}/0.660 
& \textbf{0.664}/0.606 & \textbf{0.648}/0.600 & 0.579/\underline{0.597} & \textbf{0.576}/0.570 
& \textbf{0.641}/0.597 & \textbf{0.679}/0.605 & \textbf{0.695}/0.638 & \textbf{0.813}/0.742 & \textbf{0.612}/0.582 \\
\bottomrule
\end{tabular}
}
\caption{Confidence and calibration analysis on Predictive QA (ID). Per-question-type comparison of MINT and LLM-SFT across accuracy, confidence sharpness, calibration, and correctness-separation metrics. Each entry reports MINT/LLM-SFT; bold indicates the better MINT score and underline indicates the better LLM-SFT score.}
\label{tab:confidence}
\end{table*}

\begin{table*}[ht]
\centering
\small
\resizebox{0.8\textwidth}{!}{
\begin{tabular}{lcccccc}
\toprule
& \multicolumn{3}{c}{\textbf{Answer Only}} & \multicolumn{3}{c}{\textbf{Reason and Answer}} \\
\cmidrule(lr){2-4}\cmidrule(lr){5-7}
\textbf{Dataset} & FormatFail $\downarrow$ & Acc $\uparrow$ & Macro-F1 $\uparrow$ & FormatFail $\downarrow$ & Acc $\uparrow$ & Macro-F1 $\uparrow$ \\
\midrule
Predictive QA & 0.0 & 0.832 ± 0.172 & 0.449 ± 0.147 & 0.219 & 0.799 ± 0.205 & 0.431 ± 0.146 \\
Predictive QA OOD & 0.0 & 0.696 ± 0.264 & 0.385 ± 0.141 & 0.070 & 0.636 ± 0.291 & 0.379 ± 0.158 \\
Predictive QA CoT & 0.0 & 0.830 ± 0.199 & 0.501 ± 0.244 & 0.166 & 0.803 ± 0.218 & 0.488 ± 0.242 \\
\bottomrule
\end{tabular}
}
\caption{Comparison of answer-only and reason-and-answer prompting for MINT ($h=5$) on predictive QA tasks. We report format failure rate, accuracy, and macro-F1 under deterministic decoding. Answer-only prompting uses a maximum generation length of 256 tokens, while reason-and-answer prompting uses 1024 tokens for Predictive QA and Predictive QA OOD, and 2048 tokens for Predictive QA CoT.}
\label{tab:qa_answer_reason_prompt}
\end{table*}

\begin{table*}[ht]
\centering
\small
\resizebox{0.55\textwidth}{!}{
\begin{tabular}{lcccccc}
\toprule
\textbf{Model} & Clarity $\uparrow$ & Depth $\uparrow$ & Faith $\uparrow$ & Coh $\uparrow$ & Rel $\uparrow$ & Avg $\uparrow$ \\
\midrule
LLM SFT ($h = 5$) & 3.663 & 2.689 & 3.172 & 3.166 & 3.537 & 3.246 \\
MINT ($h = 5$) & 3.553 & 2.617 & 2.960 & 3.080 & 3.457 & 3.134 \\
\bottomrule
\end{tabular}
}
\caption{LLM-as-judge evaluation of rationale quality on the Predictive QA CoT dataset. Scores are reported for clarity, depth, faithfulness, coherence, relevance, and overall average under stochastic decoding (512 max new tokens).}
\label{tab:qa_main_cot}
\end{table*}

\begin{table*}[ht]
\centering
\small
\resizebox{0.55\textwidth}{!}{
\begin{tabular}{lcccccc}
\toprule
\textbf{Model} & Faith $\uparrow$ & Cov $\uparrow$ & Spec $\uparrow$ & Coh $\uparrow$ & Conc $\uparrow$ & Avg $\uparrow$ \\
\midrule
\multicolumn{7}{l}{\textit{Structured report}} \\
LLM SFT ($h = 5$) & 4.054 & 4.409 & 3.064 & 4.707 & 3.357 & 3.916 \\
MINT ($h = 5$) & 3.480 & 4.457 & 2.856 & 4.707 & 3.794 & 3.858 \\
\midrule
\multicolumn{7}{l}{\textit{Semantic description}} \\
LLM SFT ($h = 5$) & 3.878 & 4.248 & 3.825 & 4.743 & 4.102 & 4.159 \\
MINT ($h = 5$) & 2.801 & 3.402 & 2.929 & 4.278 & 3.498 & 3.380 \\
\bottomrule
\end{tabular}
}
\caption{LLM-as-judge evaluation of caption quality. Scores are reported for faithfulness, coverage, specificity, coherence, conciseness, and overall average on structured-report and semantic-description captioning tasks under stochastic decoding (512 max new tokens).}
\label{tab:caption_main}
\end{table*}

\end{document}